\documentclass[letterpaper, 10 pt, conference]{ieeeconf} 

\IEEEoverridecommandlockouts

\usepackage{graphicx}
\usepackage{textgreek}
\usepackage{amsmath}
\usepackage{caption}
\usepackage{subcaption}
\usepackage{booktabs}
\usepackage{multirow}
\usepackage{array}

\title{\LARGE \bf
SwarmPaint: Human-Swarm Interaction for Trajectory Generation and Formation Control by DNN-based Gesture Interface 
}

\author{Valerii Serpiva$^{1}$, Ekaterina Karmanova$^{1}$, Aleksey Fedoseev$^{1}$, Stepan Perminov$^{1}$, and Dzmitry Tsetserukou$^{1}$
\thanks{$^{1}$The authors are with the Intelligent Space Robotics Laboratory, Space CREI, Skolkovo Institute of Science and Technology, Moscow, Russian Federation.
 {\tt \{valerii.serpiva, ekaterina.karmanova, aleksey.fedoseev, stepan.perminov, d.tsetserukou\}@skoltech.ru}}
}

\begin{document}

\maketitle
\thispagestyle{empty}
\pagestyle{empty}


\begin{abstract}

Teleoperation tasks with multi-agent systems have a high potential in supporting human-swarm collaborative teams in exploration and rescue operations. However, it requires an intuitive and adaptive control approach to ensure swarm stability in a cluttered and dynamically shifting environment. We propose a novel human-swarm interaction system, allowing the user to control swarm position and formation by either direct hand motion or by trajectory drawing with a hand gesture interface based on the DNN gesture recognition. 

The key technology of the SwarmPaint is the user's ability to perform various tasks with the swarm without additional devices by switching between interaction modes. Two types of interaction were proposed and developed to adjust a swarm behavior: free-form trajectory generation control and shaped formation control.

Two preliminary user studies were conducted to explore user's performance and subjective experience from human-swarm interaction through the developed control modes. The experimental results revealed a sufficient accuracy in the trajectory tracing task (mean error of 5.6 cm by gesture draw and 3.1 cm by mouse draw with the pattern of dimension 1 m by 1 m) over three evaluated trajectory patterns and up to 7.3 cm accuracy in targeting task with two target patterns of 1 m achieved by SwarmPaint interface. Moreover, the participants evaluated the trajectory drawing interface as more intuitive (12.9\%) and requiring less effort to utilize (22.7\%) than direct shape and position control by gestures, although its physical workload and failure in performance were presumed as more significant (by 9.1\% and 16.3\%, respectively). 
 
The proposed SwarmPaint technology can be potentially applied in various human-swarm scenarios, including complex environment exploration, dynamic lighting generation, and interactive drone shows, allowing users to actively participate in the swarm behavior decision on a different scale of control.

\end{abstract}

\section{Introduction}

\begin{figure}[!h]
 \includegraphics[width=1.0\linewidth]{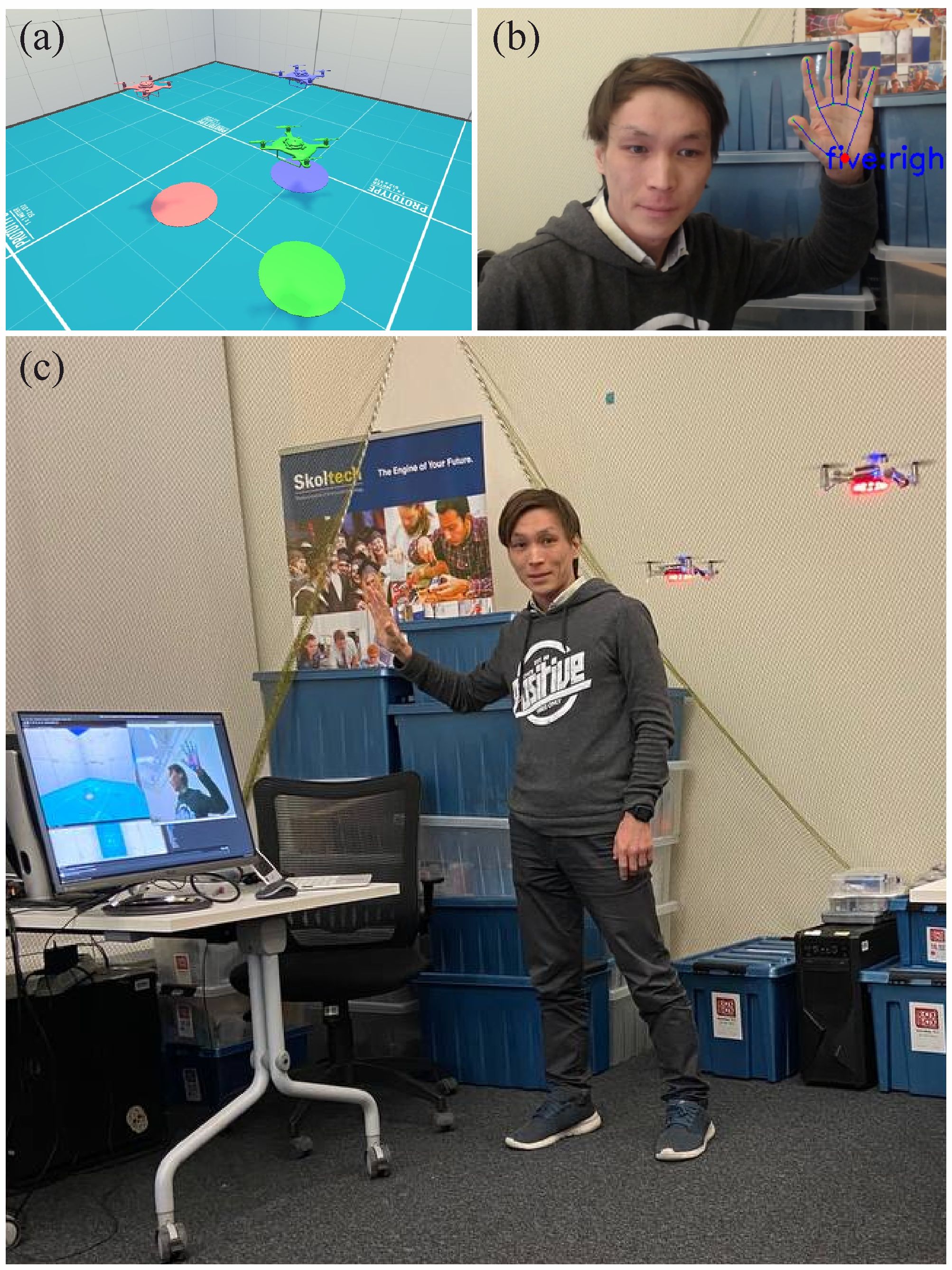}
 \caption{a) Virtual task simulation in real-time, b) gesture-based control interface, c) user controls the formation of swarm via SwarmPaint interface.}
 \label{fig:main}
\end{figure}




Nowadays, aerial drones are widely used all around the world for a variety of purposes, such as delivery, aerial videography, rescue operations, light shows, and etc. The swarm of drones can handle these tasks more efficiently than a single unit due to its scalability and task-sharing potential. However, performing complex tasks with swarm requires an experienced human teleoperator with sophisticated control interfaces, which is inefficient for the rescue teams, working in the same space as drones. Several approaches have been proposed to provide a convenient and effective interface for the swarm teleoperation.
Sarkar et al. proposed to apply a Leap Motion Controller to set the movements of a single drone by human gestures \cite{Sarkar_2016}. Vasile et al. \cite{Vasile_2011} suggested control over multi-agent systems by the developed three-module interface, including a graphical user interface module, and a pair of local and social modules for each robot in the swarm. Uryasheva et al. \cite{Uryasheva_2019} proposed a task dispatch system based on the parametric greedy algorithm for multi-drone graffiti. The brain-swarm interface implementable in multiple scenarios with visual imagery is proposed by Jeong et al. \cite{Jeong_2020}, though the robustness of such concept remained to be further validated. A potential of the brain-swarm interfaces explored by Karavas et al. \cite{Karavas_2017}, who proposed a potentially prominent approach of the hybrid input interface based on both EEG signals and an external device. 

Of all the robotic control interfaces, visual-based hand gesture recognition has been significantly improved with the emerging of the DNN approaches.
A multi-sensor system for driver's hand motion recognition was suggested by Molchanov et al. \cite{Molchanov_2015}, combining near-field radar, a color camera, and a depth camera. Liu et al. \cite{Liu_2016} introduced an interactive astronaut-robot teleoperation system integrating a data-glove with a spacesuit for the astronaut to operate through the hand gesture-based commands. In the field of mobile robotics, several researchers implemented gesture-based algorithms. Stancic et al. \cite{STANCIC_2017} proposed an effective wearable inertial-based system, that can apply hand-gesture dynamics for robotic control over high distances. The hybrid control system combining voice commands with gestures recorded via Kinect module to operate an industrial robot was presented by Kaczmarek et al. \cite{Kaczmarek_2020}. Multi-user remote control of a collaborative robot in Zoom by DNN-based gesture recognition is proposed Zakharkin et al. \cite{Zakharkin_2020}. A gesture-based drone control framework for a single aerial drone is proposed by Natarajan et al. \cite{Natarajan_2018}. 

Thus, the implementation of gesture-based systems in the field of swarm control shows a high potential due to its high accuracy and adaptability to rapidly changing environmental conditions. The next section takes under the scope the approaches, in which gestures are applied as the input data to change swarm behavior.

\section{Related Works}
 
Many gesture-based input control interfaces have been introduced in the scope of the interaction between human and robotic swarm. Nagi et al. \cite{Nagi_2014} proposed a gesture-based approach allowing the user to adjust the swarm hierarchy, dividing it into several sub-teams. A more complex control approach is suggested by Suresh et al. \cite{Suresh_2019}, where arm gestures and motions recorded by a wearable armband are applied to control a swarm shape. The gesture recognition wearable interface with multi-finger motion detection and capacitive sensing to control a single aerial drone is proposed by Montebaur et al. \cite{Montebaur_2020}. Srivastava et al. \cite{Srivastava_2020} implemented a gesture detecting sensor bracelet for an intuitive control over several drones. A tactile interface for human-swarm interaction with an impedance-controlled swarm of drones was introduced by Tsykunov et al. \cite{Tsykunov_2019}.

Alonso-Mora et al. \cite{Alonso-Mora_2015} and Kim et al. \cite{Kim_2020} suggested real-time input interfaces with complex swarm formation control. However, their approach was developed only for operation of ground mobile robots in 2D space. The wearable devices for the user's high mobility were proposed by Byun et al. \cite{Byun_2019}, suggesting epidermal tactile sensor array to achieve the direct teleoperation of the swarm by human hand. A multi-channel robotic system for human-drone swarm interaction in augmented reality is presented in Chen et al. \cite{Chen_2020}. 

Previously developed systems have achieved low time delays and high precision of the control, however, their trajectory generation capability is limited to the gesture sets and simple hand motions. To make the way we communicate with drones intuitive and intelligent, we propose a SwarmPaint system for drone light painting with DNN-based gesture recognition. Only with a single camera and developed software any not-experienced user will be capable of generating the impressive light drawings in midair. 



\section{System Overview}


The developed SwarmPaint system software consists of four modules: gesture recognition, trajectory planning, swarm control, and flight simulation module.

\begin{figure}[htbp]
 \centering
 \includegraphics[width=0.95\linewidth]{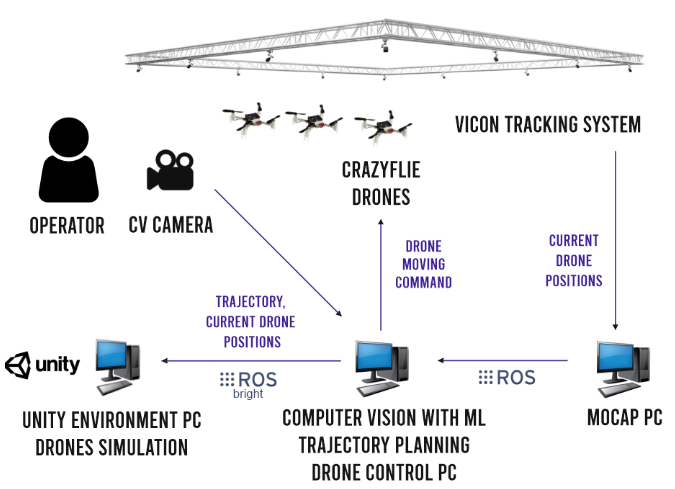}
 \caption{SwarmPaint system architecture. The three key modules perform gesture recognition, swarm behavior simulation, and real drone control, being connected via ROS environment.}
 \label{fig:node_graph}
\end{figure}

The hardware part consists of Vicon Tracking system with 12 IR cameras for drone positioning and PC with Mocap framework, a PC with CV system and drone-control framework, Logitech HD Pro Webcam C920 of @30FPS for recognizing the operator hand movements and gestures, small quadcopters Crazyflie 2.0, and PC with Unity environment for visual feedback provided to the operator (Fig. \ref{fig:node_graph}). 
Communication between all systems is performed by the ROS framework.

\subsection{Swarm guidance procedure with SwarmPaint}

\begin{figure}[!h]
 \centering
 \includegraphics[width=1\linewidth]{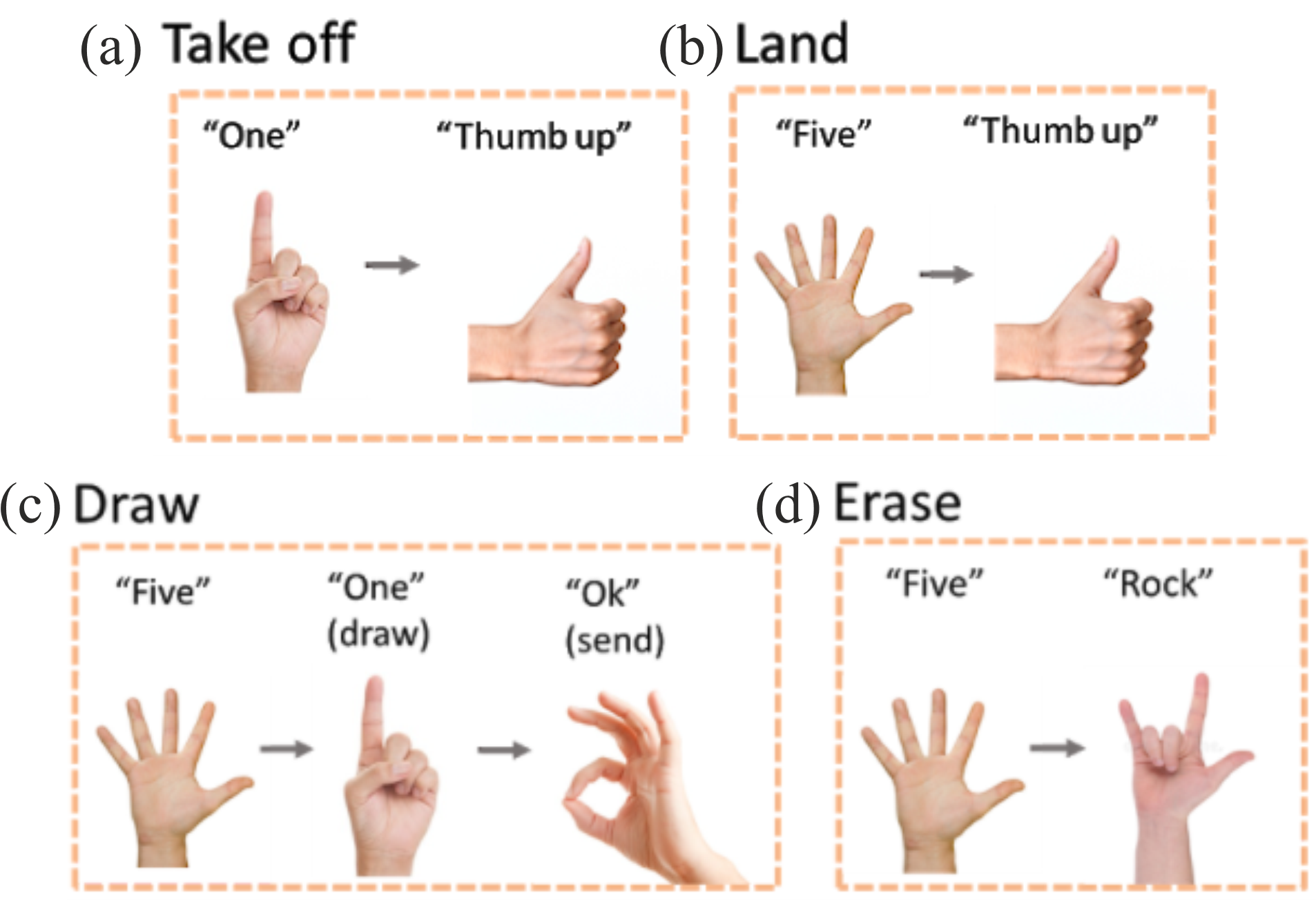}
  \caption{Example of gestures and commands in different control modes. The colored box frames indicate the control mode corresponding to the displayed command. The blue, green, red, and yellow frames define a swarm control, free shape control, trajectory drawing, basic control commands, respectively.}
 \label{fig:commands}
\end{figure}

Before deploying the drone swarm, the operator positions himself in front of the web camera, which sends the captured footage to the gesture recognition module. As soon as the SwarmPaint is activated, the module starts to recognize the operator's position and hand gestures, awaiting the ‘Take off’ command to deploy the swarm (Fig. \ref{fig:commands}a).

After the drones are taken off, the user selects the control mode, in which they intend to manipulate the swarm. In trajectory drawing mode, the gesture recognition module generates a trajectory drawn by the operator with gestures. The developed SwarmPaint interface allows the operator to draw and erase the trajectory to achieve the desired result (Fig. \ref{fig:commands}). After that, the drawn path is processed by the trajectory planning module to make it suitable for the swarm. ROS framework then sends the processed trajectory simultaneously to the swarm control module and the flight simulation module. Finally, the drones are driven by the swarm control module to follow the received trajectory.

While performed by the swarm of drones, all the trajectories are being displayed at the SwarmPaint interface screen in real-time. Additionally, the operator can follow the current position of each drone in the Unity simulation environment. 

In the formation-control mode, the user may select the shape of the swarm from the limited set and apply it to the real drones via corresponding gestures. In this case, the distance between the hand's fingers defines the relative distance between swarm units, the inclination of the hand controls the rotation of the swarm, while the hand position on the screen defines the position of formation's center of mass  (CoM) in the scaled 3D space of the room.

Drones complete their work at the command ‘Land' performed by the operator with the appropriate gesture (Fig. \ref{fig:commands}b).

\subsection{Gesture Recognition}

We developed a hand-tracking system with DNN to achieve an intuitive way of drone trajectory generation. The Mediapipe framework was used to implement the hand-tracking modules. It provides high-fidelity tracking of the hand by employing Machine Learning (ML) to infer 21 key points of a human hand per a single captured frame.

\begin{figure}[h]
 \centering
 \includegraphics[width=0.9\linewidth]{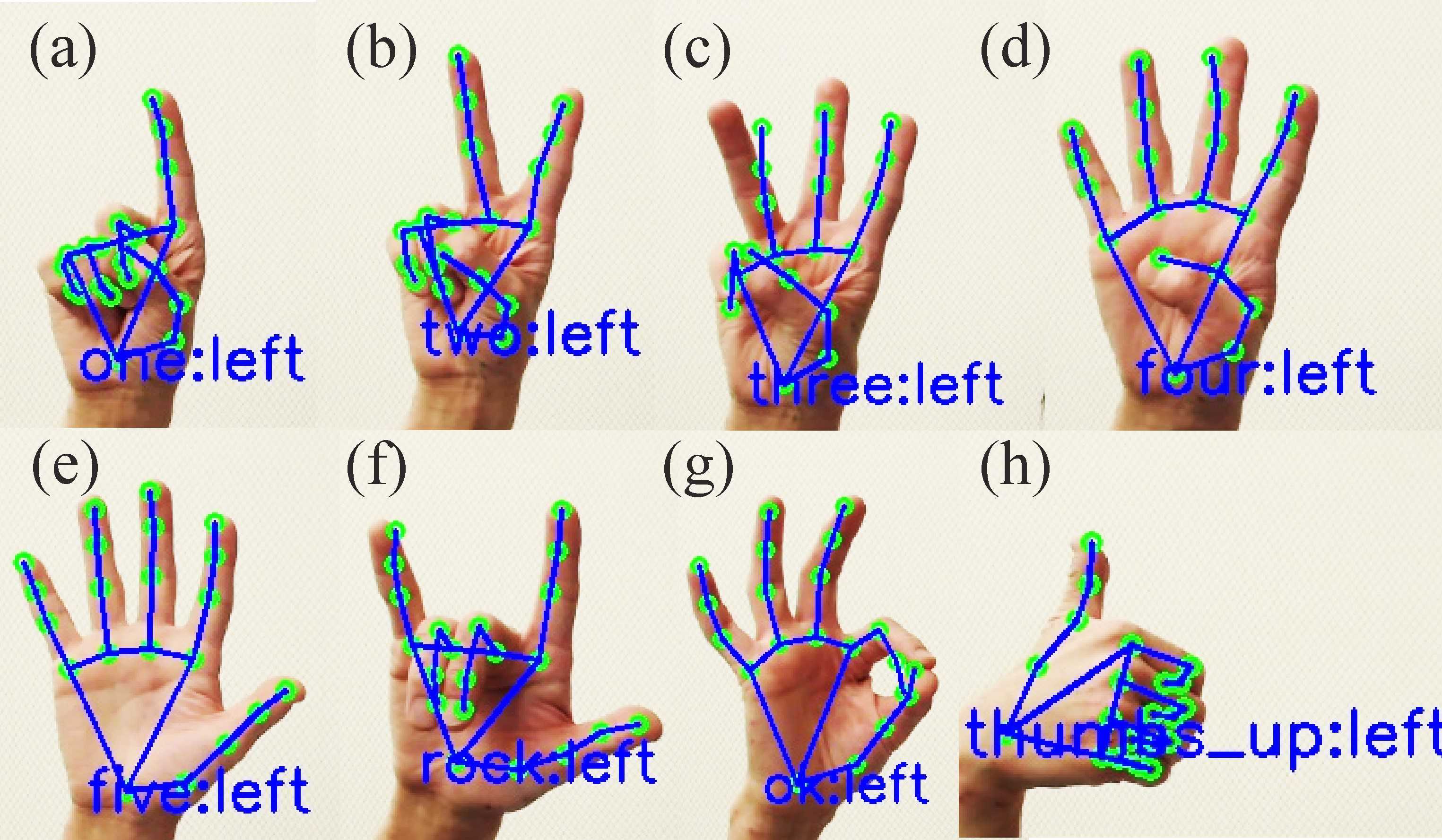}
 \caption{Recognition of eight input gestures, selected for DNN training and swarm control.}
 \label{fig:gestures}
\end{figure}

For convenient drone control we propose 8 gestures: “one", “two", “three", “four", “five", “okay", “rock", “thumbs up", which are displayed in Fig. \ref{fig:gestures}.

\begin{figure}[htbp]
 \centering
 \includegraphics[width=0.90\linewidth]{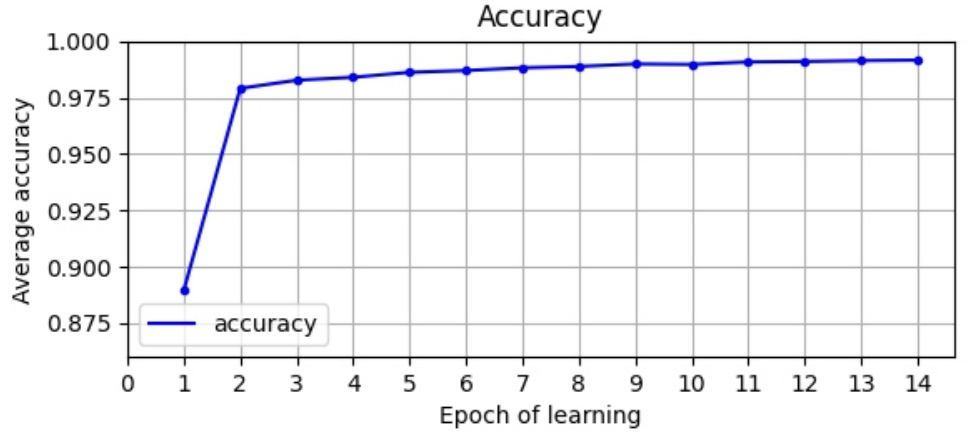}
 \caption{The classification accuracy of the developed gesture recognition system.}
 \label{fig:acc_loss}
\end{figure}

A gesture dataset for the training model was recorded from five participants. The gesture dataset consists of 8000 arrays with coordinates of 21 key points of a human hand: 1000 per gesture (200 per person). We then applied normalized landmarks, i.e., angles between the joints and pairwise landmark distances as features to predict the gesture class.
It allowed us to achieve an accuracy of 99.75\% when performing validation on a test set (Fig. \ref{fig:acc_loss}).

\subsection{Trajectory Processing}

\begin{figure}[htbp]
 \centering
 \includegraphics[width=0.95\linewidth]{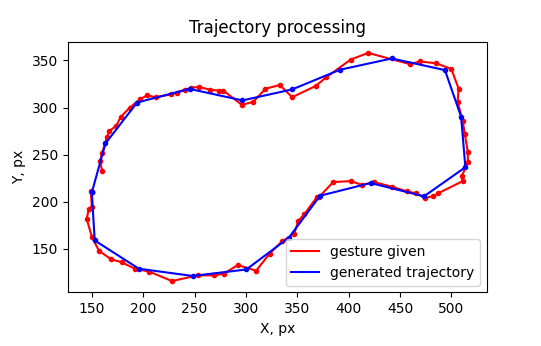}
 \caption{Hand trajectory normalization. Trajectory recorded from the gesture movement (red line). Trajectory with filtration and linear interpolation (blue line).}
 \label{fig:traj_processing}
\end{figure}

To ensure a smooth flight of the drone, a trajectory with equidistant flight coordinates is required. The drawn trajectory may be uneven or contain unevenly distributed coordinates. Therefore, the trajectory processing module smooths the drawn trajectory with an alpha-beta filter (filtration coefficient of 0.7) and then interpolates it uniformly (Fig. \ref{fig:traj_processing}). After that, the trajectory coordinates are transformed from the SwarmPaint interface screen (pixels) to the flight zone coordinate system (meters). 
The coordinates of the generated trajectory are sequentially transferred to the drone control system using the ROS framework. The time intervals between each coordinate transfer depend on the distance between the coordinates and the flight speed of the drone.

The hand coordinates $h_x, h_y$, and palm size $h_s$ are transformed from the screen pixels to the drone coordinates $d_x, d_y, d_z$ in meters as shown in Eq. (\ref{eq: transform}):
\begin{equation}
 \begin{cases}
 d_x = L_x \left( \frac{h_x}{F_x} - \frac{1}{2} \right) \\
 d_y = L_y \left( \frac{h_s}{h_{s max}} - \frac{1}{2} \right) \\
 d_z = L_z \frac{h_y}{F_y},
 \end{cases}
\label{eq: transform}
\end{equation}
where $L$ is the length of the drone flight zone in meters, $F$ is the screen resolution in pixels, and $h_{s max}$ is the size range of the operator's palm in pixels determined in the preliminary calibration.

\subsection{Swarm Control Algorithm}

The potential field approach was adjusted and applied to UAVs for robust path planning and collision avoidance between the swarm units. The basic principle of this method lies in the modeled force field, which is composed of two opposing forces, i.e., attractive force and repulsive force.
The attractive force pulls the UAV to the desired position, located on the drawn and processed trajectory, while the repulsive force repels the UAV. The repulsive force centers are located on the obstacle surfaces and on the other UAVs.

The applied attractive potential $U_a$ is described as:

\begin{equation}
 U_a(q) = \frac{1}{2} k_a \left( q - q_g \right)^2 ,
 \label{eq: potencial}
 \end{equation}
where $q$ is the current position of the drone, $q_g$ is the goal position which the drone has reached, and $k_a$ is the attraction constant.

The repulsive potential function $U_r$, which generated by other UAV can be written as:

\begin{equation}
 U_r(q)=\begin{cases}
 \frac{1}{2} k_r \left( \frac{1}{q} - \frac{1}{r_0} \right)^2 q^2  & \text{$q<=r_0$} \\
 0 & \text{$q>r_0$}
 \end{cases},
 \label{eq: repulsive}
 \end{equation}
 where $q$ is the distance between current UAV and an obstacle, $r_0$ is the limit distance which is influenced by the repulsive potential and $k_r$ is the repulsive constant. 

Therefore, total potential function $U$ which lead UAV to the target point is defined by: 

\begin{equation}
 U(q) = U_a(q) + \sum U_r(q).
 \label{eq: total}
\end{equation}

The coefficients of the fields of attraction and repulsion were calculated experimentally in the simulation when two groups of drones flew along an intersecting trajectory. With the values of the coefficients $k_r = 30$, $k_a = 1$ and safe distance $r_o = 0.45$ m, drones get close to the minimum distance $r_{min} = 0.36$ m keeping a smooth movement along the trajectory.

\subsection{Swarm Simulation}
\begin{figure}[htbp]
 \centering
 \includegraphics[width=0.9\linewidth]{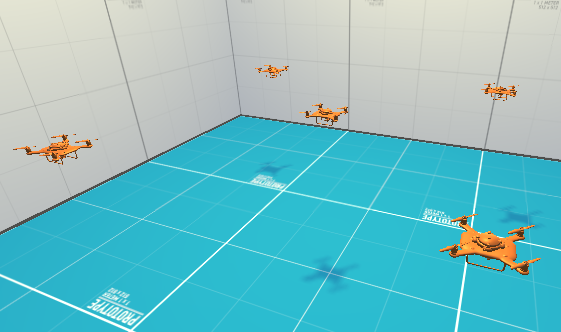}
 \caption{Drone performing square formation in the Unity simulation environment.}
 \label{fig:unity}
\end{figure}

To visualize the swarm performance before task execution, we have developed a simulation environment in Unity Engine. The environment supports both swarm simulation and visualization of the real swarm, allowing a user to control its behavior from the remote area. To simulate flight dynamics of quadrotors, supporting the control of drone position and orientation in space, an approximation by thrust forces was implemented. In our simulation, we introduced downward thrust force along the drone's CoM to approximate the combined thrust of propellers. 
The drone's position is controlled by rotation angle variables around roll and pitch axes limited heuristically to 20 deg. 
A discrete PID control was applied as a control algorithm, whereas the output values of tilt angle around roll and pitch were calculated from the target position. The swarm accomplished a mission and potential field behaviors between agents through the trajectory following and formation control (Fig. \ref{fig:unity}).

\section{Experimental Evaluation}
We conducted a two-part user evaluation to determine the extent to which the developed teleoperation modes increase the user's ability for real-time swarm control. For this, we analyzed the performance of our two proposed interactions through gestures: trajectory drawing with determining the accuracy of path generation and landing task by free-form behavior with formation control. Each test contained both the quantifiable parameters to access the dexterity of swarm teleoperation and the survey with a 7-point Likert scale.

\subsubsection*{Participants}
We invited 7 participants aged 22 to 28 years (mean = 24.7, std = 1.98) to test SwarmPaint system. 14.3\% of them have never interacted with drones before, 28.6\% regularly deal with drones, almost 87\% of participants were familiar with CV-based systems or had some experience with gesture recognition.

\subsection{Experiment on Trajectory Generation}

\subsubsection{Tracing Accuracy Task} To evaluate the performance of the proposed control interface, we collected trajectories drawn by gestures and computer mouse from the participants, after which the real drone has followed the drawn trajectory.
As a reference point in determining the ease and accuracy of the proposed method, drawing by the mouse was chosen as convenient to the user.

\begin{figure}[htbp]
 \centering
 \includegraphics[width=0.8\linewidth]{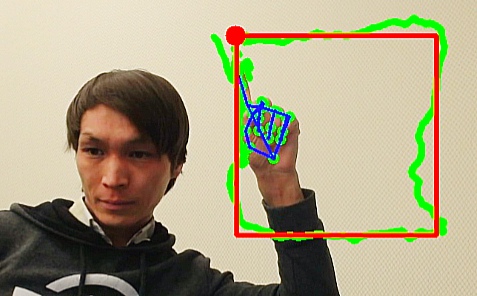}
 \caption{View of the user screen. Hand drawing trajectory (green line) and given trajectory “square" (red line).}
 \label{fig:drawing_process}
\end{figure}

 In the first round, each participant drew given trajectory by hand five times  (Fig. \ref{fig:drawing_process}). In the second round, the same trajectories were drawn by a computer mouse.

After that, three best attempts were chosen to evaluate performance and compare methods. 

\subsubsection{User Experience Evaluation} After performing the one-time trajectory drawing with the real drone, all respondents were asked to evaluate two control interfaces with a Likert scale based on the unweighted NASA Task Load Index (TLX). Control interfaces such as SwarmPaint system (using hand gestures) and computer mouse were considered.
An extra “intuitiveness" parameter was introduced in the survey, being an essential criterion in the teleoperation tasks to provide adjustable control over the swarm behavior in real-time.
Therefore, the participants provided feedback on seven questions: 
 
 \begin{itemize}
 \item Mental Demand: How much perceptual and mental activity was required for the swarm control (e.g. measuring, deciding, and etc.)? Was this task simple or complex? (Low - High)
 
 \item Physical Demand: How much physical activity was required during the teleoperation? (Low - High)
 
 \item Temporal Demand: How much time pressure did you feel due to the response time at which the drones performed their task? (Low - High)
 
  \item Overall Performance: How successful do you think you accomplished the task? How satisfied are you with your performance? (Perfect - Failure)

 \item Effort: Was it hard for you to fulfill the task (mentally and physically) with your level of success?  (Low - High)

 \item Frustration Level: Did you feel stressed or annoyed during the task or did you feel complacent and relaxed? (Low - High)
 
 \item Intuitiveness: How natural was the experience? How intuitive did you find gesture combinations for drones control? (Intuitive - Artificial)
 \end{itemize}

\subsection{Experimental Results}

\subsubsection{Tracing Accuracy Task} 
\begin{figure}[!htbp]
 \centering
 \begin{subfigure}{1\linewidth}
 \includegraphics[width=\textwidth]{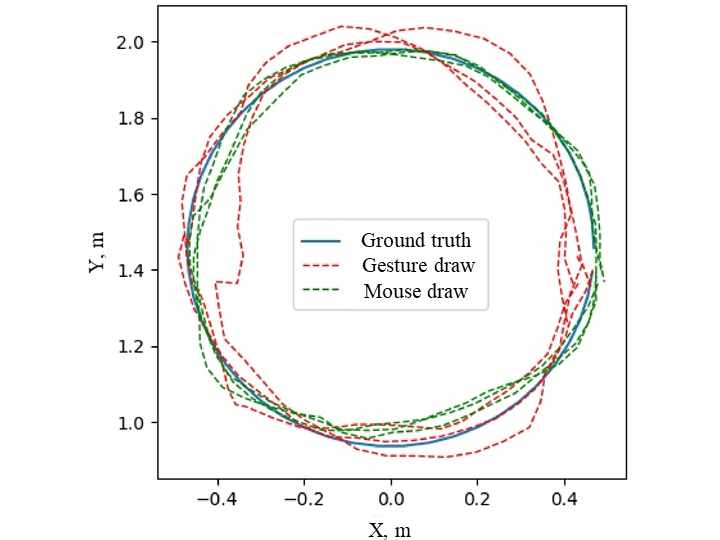}
 \caption{Circle trajectory.}
 \label{fig:circle}
 \end{subfigure}
\hskip2em
\begin{subfigure}{1\linewidth}
\includegraphics[width=\textwidth]{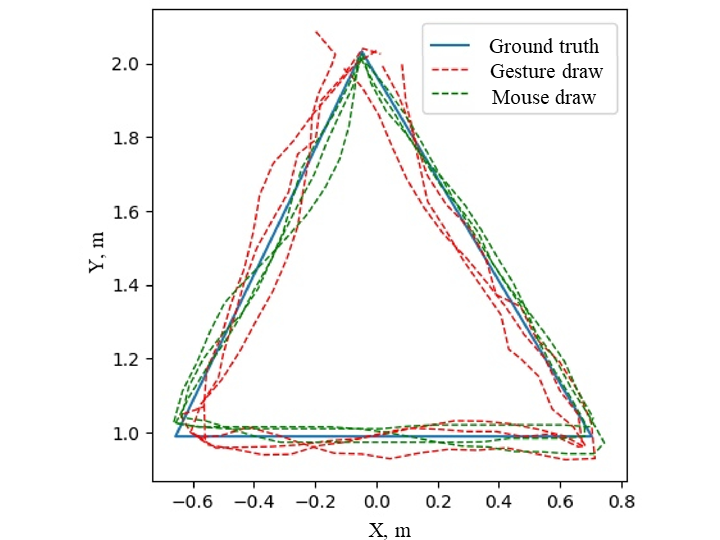}
\caption{Triangle trajectory.}
\label{fig:triangle}
\end{subfigure}
\caption{
a) Circle and b) triangle trajectories drawn by the hand gestures (red dashed line) and the mouse (green dashed line).}
\label{fig:plots}
\end{figure}

Fig. \ref{fig:plots} shows two out of three ground truth paths (blue lines), where for each one, the user traces the path several times using gestures (red dashed line) and the mouse (green dashed line).
The experimental results are presented in Table \ref{tab:error}. 
Max errors, mean errors, and RMSE (Root Mean Square Errors) were considered for the result evaluation.


\begin{table}[htbp]
\small
\caption{Comparison experiment for trajectories drawn by the hand (H) and mouse (M)}
\newcolumntype{A}{m{0.03\linewidth}}
\newcolumntype{B}{m{0.23\linewidth}}

\begin{tabular}{|B|AAAAAA|}
\hline
 & & & Trajectories & & & \\ \cline{2-7} 
 & Square & \multicolumn{1}{c|}{} & Circle & \multicolumn{1}{c|}{} & Triangle & \\ \cline{2-7} 
 & \multicolumn{1}{c|}{H} & \multicolumn{1}{c|}{M} & \multicolumn{1}{c|}{H} & \multicolumn{1}{c|}{M} & \multicolumn{1}{c|}{H} & \multicolumn{1}{c|}{M} \\ \hline
Max error, m & \multicolumn{1}{c|}{0.19} & \multicolumn{1}{c|}{0.11} & \multicolumn{1}{c|}{0.17} & \multicolumn{1}{c|}{0.08} & \multicolumn{1}{c|}{0.13} & \multicolumn{1}{c|}{0.06} \\ \hline
Mean error, m & \multicolumn{1}{c|}{0.06} & \multicolumn{1}{c|}{0.04} & \multicolumn{1}{c|}{0.06} & \multicolumn{1}{c|}{0.03} & \multicolumn{1}{c|}{0.04} & \multicolumn{1}{c|}{0.02} \\ \hline
RMSE, m & \multicolumn{1}{c|}{0.08} & \multicolumn{1}{c|}{0.05} & \multicolumn{1}{c|}{0.08} & \multicolumn{1}{c|}{0.04} & \multicolumn{1}{c|}{0.05} & \multicolumn{1}{c|}{0.03} \\ \hline
Time, sec & \multicolumn{1}{c|}{15.50} & \multicolumn{1}{c|}{5.52} & \multicolumn{1}{c|}{13.47} & \multicolumn{1}{c|}{4.89} & \multicolumn{1}{c|}{12.04} & \multicolumn{1}{c|}{4.50} \\ \hline
\end{tabular}
\label{tab:error}
\end{table}

The purpose of the study was to compare
the accuracy in two trajectory generation methods (gesture-drawn and mouse-drawn trajectories). 
The overall mean error equals 5.6 cm (95\% confidence interval [CI], 4.6 cm to 6.6 cm) for the gesture-drown trajectories and 3.1 cm (95\% CI, 2.7 cm to 3.5 cm) for the mouse-given trajectories.
The ANOVA results showed a statistically significant difference between the user's interaction with trajectory patterns: $F$ = 4.006, $p$-value = 0.025 $\textless$ 0.05.
On average the trajectory, generated with gestures is 2.5 cm farther from the ‘Ground truth' than the one drawn with a computer mouse. That difference was hypothesized to occur due to three reasons:
\begin{itemize}

 \item The hand is trembling while drawing the trajectory in midair. That issue might be neutralized either by data post-processing or by the implementation of a hand-supporting device.
 \item The lack of tangible feedback from the control device. That issue might be managed either by improving data post-processing or by integration of a haptic device, which might provide the required feedback.
 \item The noise caused by system delays. It can be reduced by optimizing the drone internal control algorithm. 
\end{itemize}

\subsubsection{User Experience Evaluation} 
\begin{figure}[htbp]
 \centering
 \includegraphics[width=1\linewidth]{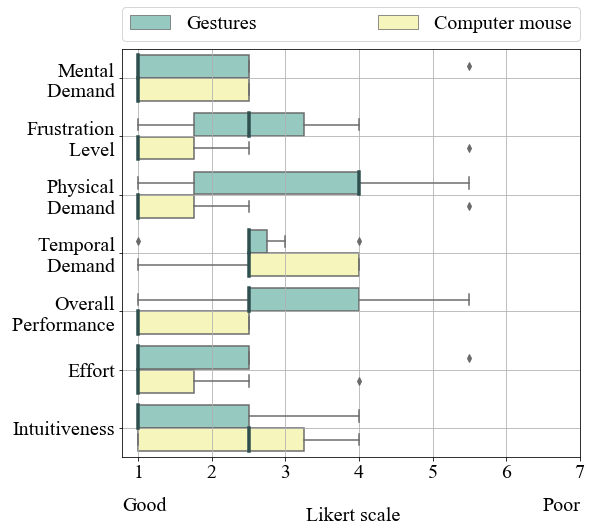}
 \caption{Subjective feedback on 7-point Likert scale: Hand gesture and Computer mouse control interface comparison.}
 \label{fig:likert_scale_compare}
\end{figure}

The results of evaluation of both control interfaces, such as SwarmPaint and computer mouse, are provided in the Fig. \ref{fig:likert_scale_compare}. 


According to the results, the trajectory tracing task was fairly achievable for both control interfaces. However, the stress level achieved by participants during the task was higher in the case of SwarmPaint (median = 2.5) because of its novelty in comparison with the well-known computer mouse (median = 1.0). The level of required physical activity was also mentioned by participants to be higher for SwarmPaint (median = 4.0 vs 1.0). The reason for high physical demand is most likely caused by the posture remaining requirement when the user had to sit in front of the monitor with the raised hand instead of making regular mouse movements. Surprisingly, some participants felt more time pressure while operating the mouse. Despite this fact, they were less satisfied while tracing the trajectories by SwarmPaint (median = 2.5 vs 1.0), as it provided less accuracy according to their own perception. 

Some participants mentioned that it was difficult for them to use SwarmPaint instead of the mouse. However, the Effort index was almost equally high for both control interfaces (median = 1.0). Finally, SwarmPaint intuitiveness level appeared to be confidently superior to the one provided by the computer mouse (median = 1.0 vs 2.5). This result was achieved because SwarmPaint allowed controlling the cursor on the monitor and, therefore, tracing the trajectories directly by hand, while the mouse acted as an intermediary between the hand and cursor.

\subsection{Experiment on the Swarm Formation Control}
 

\subsubsection{Targeting Accuracy Task} 
\begin{figure}[htbp]
 \centering
 \includegraphics[width=1\linewidth]{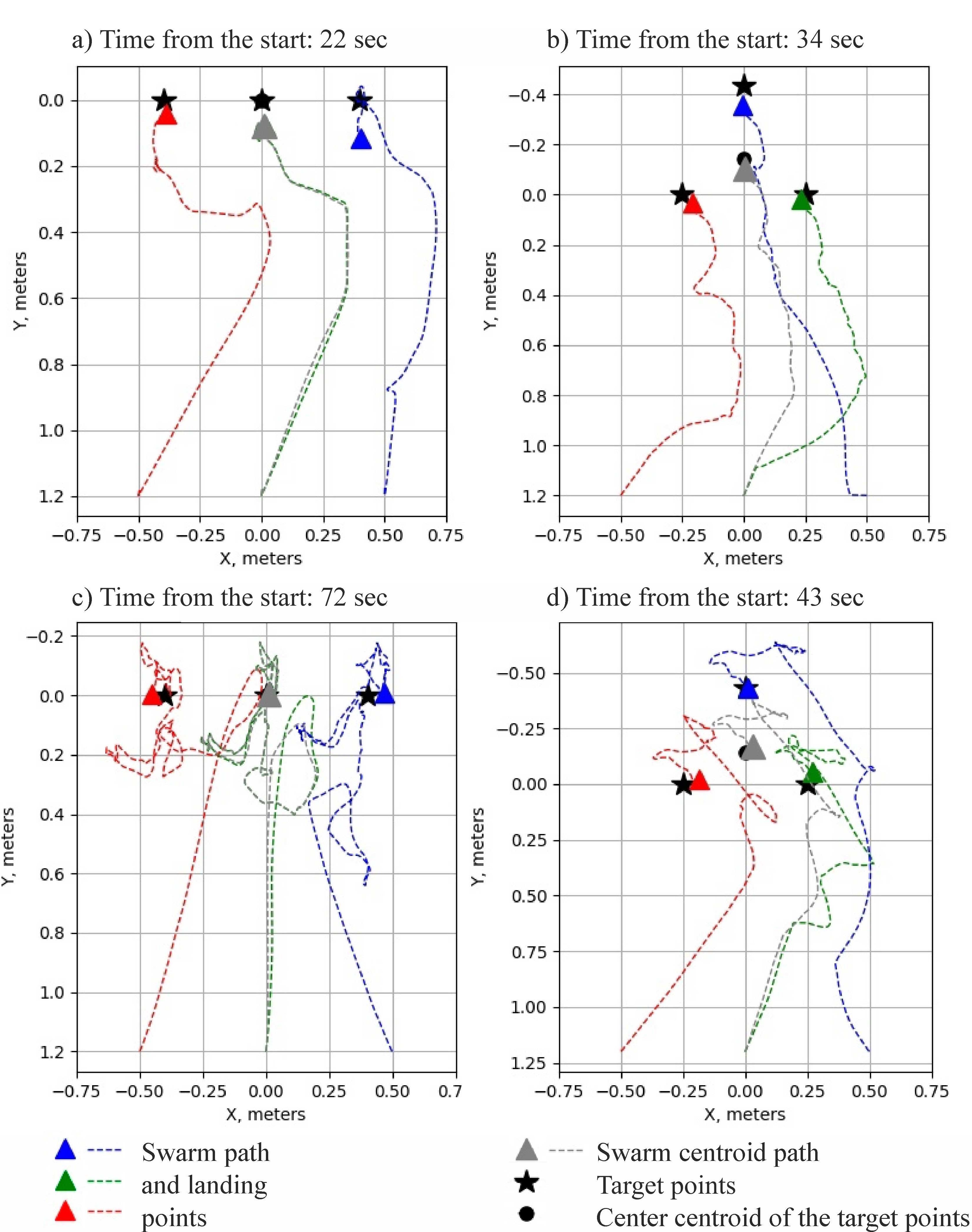}
 \caption{A top view of trajectories of drones and centroid while being manipulated toward the targets by SwarmPaint interface. Left column of two pictures (a, c) represents linear formation of three drones. Right column (b, d) represents the triangular shape of the swarm. Gray line is
the trajectory of the swarm centroid, controlled by the palm motion and rotation. Dashed lines are actual drones trajectories, controlled by the distance between fingers. Stars represent the desired targets.}
 \label{fig:swarm_formation_contro}
\end{figure}

The task was performed in the simulation environment to neglect all errors caused by real drone inertia and hovering behavior, focusing on a high-level control approach. During this mission, the participants controlled a swarm of drones with three units to navigate and land them on targets, located in various patterns on the floor. Each swarm unit and the corresponding target were highlighted by the same color (Fig. \ref{fig:swarm_formation_contro}). Two patterns were selected for this task: a line and a triangle, allowing evaluation of both linear and volumetric swarm formation.

Each trial was initiated with the swarm deployment described in section III.A. Participants first selected the formation of the pattern with a gesture, and then, with a motions of their hand, positioned the swarm as accurately as possible to reach the targets. Both the line and triangle patterns required the participants to rotate the swarm around its CoM and to maintain a certain distance between the units with finger motions. After reaching the goals, the participants showed a “thumb up" gesture, which completed the trial.

At the beginning of the experiment, the procedure and system equipment were introduced to the participants. Each participant has been given two trials of the task scenario, followed by the evaluated performance. After every launch, the experiment duration and simulated trajectories from the Unity environment system were recorded.

\subsubsection{User Experience Evaluation} All respondents were asked to evaluate the SwarmPaint formation control mode with a Likert scale based on the unweighted NASA Task Load Index (TLX) (1-7) on 7 issues: Mental Demand, Physical Demand, Temporal Demand, Overall Performance, Effort, Frustration Level, and Intuitiveness.

\subsection{Experimental Results}
\subsubsection{Targeting Accuracy Task}

The purpose of the study was to estimate if there is a dependence between trajectory mean error and target pattern (Land structure). For this purpose, we calculate the average error for each drone and the error of swarm's geometrical centroid (Table \ref{tab:Averaged experiment 2 results}).


The overall mean error of swarm center point landing was 4.74 cm (95\% CI, 2.91 cm to 6.56 cm) with max value for the drone displacement of 21.60 cm. The mean error for each drone were 6.72 cm (95\% CI, 4.39 cm to 9.05 cm), 7.34 cm (95\% CI, 4.05 cm to 10.6 cm), and 8.55 cm (95\% CI, 5.43 cm to 11.67 cm). While the experiment was conducted in a simulated virtual environment without hardware issues, we can say that the drone-error dependence proved that the drone position in a swarm significantly affects its performance.

The ANOVA results showed that there is no statistically significant difference between the evaluated patterns ($F$ = 0.05, $p$-value = 0.82 $\textgreater$ 0.05). To investigate the case of such overall results, we ran a one-way ANOVA analysis between each drone's mean errors. We found a significant main effect on the error of the two out of three swarm units ($F$ = 5.3, $p$-value = 0.04 $\textless$ 0.05) and 
($F$ = 4.20, $p$-value = 0.042 $\textless$ 0.05), suggesting that user precision in targeting might depend in most cases on the performed pattern, however, there is a presence of repetitive error in drone positioning, which is independent of the swarm formation shape. 

\begin{table}[htbp]
\fontsize{10}{11}\selectfont
\caption{Targeting error with two swarm patterns}

\centering
\begin{tabular}{|l|*{2}{c}}
\hline
 & \multicolumn{2}{c|}{Target pattern} \\ \cline{2-3} 
\makebox[30mm]{ } &\multicolumn{1}{c|}{Line }&\multicolumn{1}{c|}{Triangle}\\
\hline
Drone 1 mean error, cm & \multicolumn{1}{c|}{8.9} & \multicolumn{1}{c|}{8.9} \\
\hline
Drone 2 mean error, cm & \multicolumn{1}{c|}{4.5} & \multicolumn{1}{c|}{10.1} \\
\hline
Drone 3 mean error, cm &\multicolumn{1}{c|}{8.5} & \multicolumn{1}{c|}{8.6} \\
\hline
Formation mean error, cm & \multicolumn{1}{c|}{7.3} & \multicolumn{1}{c|}{7.8} \\
\hline
Mean formation centroid error, cm & \multicolumn{1}{c|}{4.3} & \multicolumn{1}{c|}{5.2} \\
\hline
Max formation centroid error, cm & \multicolumn{1}{c|}{11.3} & \multicolumn{1}{c|}{8.8} \\
\hline
Time, sec & \multicolumn{1}{c|}{41} & \multicolumn{1}{c|}{44} \\
\hline
\end{tabular}
 
\label{tab:Averaged experiment 2 results}
\end{table}

\subsubsection{User Experience Evaluation}

\begin{figure}[htbp]
 \centering
 \includegraphics[width=1\linewidth]{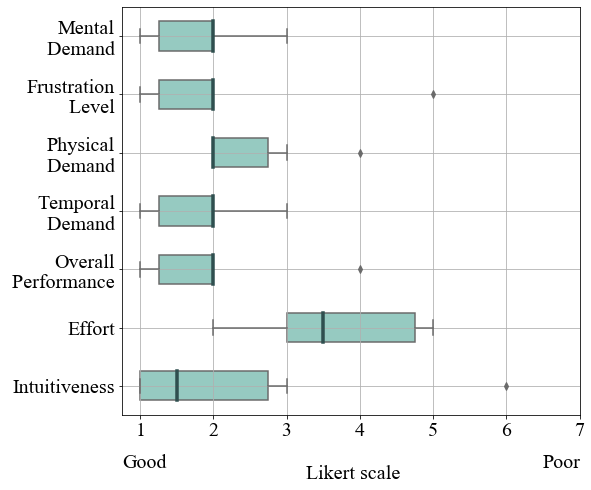}
 \caption{Subjective feedback on 7-point Likert scale: Swarm formation control.}
 \label{fig:likert_scale_swarm}
\end{figure}

The results of subjective evaluation survey are presented in Fig. \ref{fig:likert_scale_swarm}.


The participants estimate the task as being easy to perform, therefore, the evaluation criteria showed better results on the Mental Demand issue (mean = 1.83, (95\% CI, 1.04 to 2.62)), while the previous experiment was considered to be slightly more demanding (mean = 2.07, (95\% CI, 0.53 to 2.62)).
This experiment requires less static hand positions. Thus, the operator does not feel tired for a long time during the control procedure, which results in Physical Demand in free-form mode (mean 2.50, (95\% CI, 1.62 to 3.38)), being evaluated by 0.64 points lower than in the trajectory drawing mode (mean = 3.14, (95\% CI, 1.57 to 4.72)). 

The Overall Performance (mean = 2.0, 95\% CI (0.85, 3.15), SD = 1.09) was proved to be better than with the trajectory generation experience (mean = 3.15 95\% CI (1.79, 4.5)), participants were more satisfied with the result using gestures to control the configuration of the swarm rather than generating patterns in the midair.

We conducted a chi-square analysis, based on the frequency of answers in each category.
The results showed the independence of chosen parameters. The min $p$-value appeared for the Frustration level and Physical demand parameters ($\tilde{\chi}^2$ = 8.25, $p$-value= 0.08 $\textgreater$ 0.05). Surprisingly, the highest correlation was revealed between Intuitiveness and Temporal Demand parameters ($\tilde{\chi}^2$ = 4.67, $p$-value= 0.59 $\textgreater$ 0.05), though at the beginning of the experiment we assumed Intuitiveness criteria being more likely dependent on Effort ($\tilde{\chi}^2$ = 14.0, $p$-value= 0.12 $\textgreater$ 0.05) or Frustration level ($\tilde{\chi}^2$ = 6.67, $p$-value= 0.35 $\textgreater$ 0.05) parameters.






\section{Conclusions and Future Work}

We have developed a novel swarm control interface in which an operator leads the swarm by path drawing and formation control with the DNN-based gesture interface and trajectory generation systems. The proposed interaction scenarios utilize the advantages of sequential gesture control to preserve the formation control while performing the intuitive drawing of swarm trajectories, inapplicable by the direct teleoperation. Thus, SwarmPaint delivers a convenient and intuitive toolkit to the user without any wearable devices, achieving high accuracy and variety of the swarm behavior. The developed system allowed the participants to achieve sufficient accuracy in trajectory generation (average error by 5.6 cm, max error by 9 cm higher than corresponding values during the mouse input) and targeting task (mean error of 7.3 cm on a linear and 7.8 cm on a triangular pattern).


As to the user experience evaluation, the participants generally evaluated SwarmPaint as providing a greater level of intuitiveness than commonly used control interfaces, i.e., the computer mouse (median = 1.0 vs 2.5 assesses by 7-point Likert scale). Additionally, the targeting task evaluation showed considerably higher score for Mental Demand, Physical Demand, and Overall Performance (mean = 1.83, 2.50, 2.00, respectively) in comparison with the trajectory tracing task (mean = 2.07, 3.14, 3.15, respectively).

The proposed SwarmPaint system can potentially have a big impact on movie shooting to achieve desirable lighting conditions with the swarm of spotlights controlled by the operator's gestures. Additionally, it can be used in a new generation of light show where each spectator will be capable of controlling the drone display, e.g., navigate the plane, launch the rocket, or even draw the rainbow in the night sky.





\section*{Acknowledgment}
The reported study was funded by RFBR and CNRS, project number 21-58-15006.

\bibliographystyle{IEEEtran}
\bibliography{sample-base}
\end{document}